\documentclass[sn-mathphys,Numbered]{sn-jnl}

\usepackage{graphicx}%
\usepackage{multirow}%
\usepackage{amsmath,amssymb,amsfonts}%
\usepackage{amsthm}%
\usepackage{mathrsfs}%
\usepackage[title]{appendix}%
\usepackage{xcolor}%
\usepackage{textcomp}%
\usepackage{manyfoot}%
\usepackage{booktabs}%
\usepackage{algorithm}%
\usepackage{algorithmicx}%
\usepackage{algpseudocode}%
\usepackage{listings}%
\usepackage{arydshln}

\theoremstyle{thmstyleone}%
\theoremstyle{thmstyletwo}%

\theoremstyle{thmstylethree}%

\newcommand{\new}[1]{\textcolor{black}{#1}}

\begin{document}

\title[Article Title]{Shift-Reduce Task-Oriented Semantic Parsing with Stack-Transformers}


\author{\fnm{Daniel} \sur{Fern\'andez-Gonz\'alez}}\email{danifg@uvigo.gal}

\affil{\orgname{Universidade de Vigo}, \orgdiv{Departamento de Informática}, \orgaddress{\street{Campus As Lagoas s/n}, \postcode{32004}, \city{Ourense}, \country{Spain}}}

\abstract{
\textbf{Introduction:} Intelligent voice assistants, such as Apple Siri and Amazon Alexa, are widely used nowadays. These task-oriented dialogue systems require a semantic parsing module in order to process user utterances and understand the action to be performed.
This semantic parsing component was initially implemented by rule-based or statistical slot-filling approaches for processing simple queries; however, the appearance of more complex utterances demanded
the application of shift-reduce parsers or sequence-to-sequence models. Although shift-reduce approaches were initially considered the most promising option, the emergence of sequence-to-sequence neural systems has propelled them to the forefront as the highest-performing method for this particular task. In this article, we advance the research on shift-reduce semantic parsing for task-oriented dialogue.

\textbf{Methods:}  We implement novel shift-reduce parsers that rely on Stack-Transformers.
This framework allows to adequately model transition systems on the Transformer neural architecture, notably boosting shift-reduce parsing performance. Furthermore, our approach goes beyond the conventional top-down algorithm: we incorporate alternative bottom-up and in-order transition systems derived from constituency parsing into the realm of task-oriented parsing.

\textbf{Results:} We extensively test our approach 
on multiple domains from
the Facebook TOP benchmark,
improving over existing shift-reduce parsers and state-of-the-art sequence-to-sequence models in both high-resource and low-resource settings. We also empirically prove that the in-order algorithm substantially outperforms the commonly-used top-down strategy.

\textbf{Conclusion:}  Through the creation of innovative transition systems and harnessing the capabilities of a robust neural architecture, our study showcases the superiority of shift-reduce parsers over leading sequence-to-sequence methods on the main benchmark.

\textbf{This version of the article has been accepted for publication in Cognitive Computation after peer review, but is not the Version of Record and does not reflect post-acceptance improvements, or any corrections. The Version of Record is available online at: \url{https://doi.org/10.1007/s12559-024-10339-4}.}
}




\maketitle

\section{Introduction}
The research community and industry have directed significant attention towards the advancement of intelligent personal assistants such as Apple Siri, Amazon Alexa, and Google Assistant. These systems, known as \textit{task-oriented dialogue} systems, streamline task completion and information retrieval via natural language interactions within defined domains such as media playback, weather inquiries, or restaurant reservations. The increasing adoption of these voice assistants by users has not only transformed individuals' lives but also impacted real-world businesses.

\new{Humans effortlessly understand language, deriving meaning from sentences and extracting relevant information. \textit{Semantic parsing} attempts to emulate this process by understanding the meaning of natural language expressions and translating them into a structured representation that can be interpreted by computational systems. Therefore, a crucial component of any voice assistant is a \textit{semantic parser} in charge of natural language understanding. 
  Its purpose is to process user dialogue by converting each input utterance into an unequivocal task representation understandable and executable by a machine.
Specifically, these parsers identify the user's requested task intent (\textit{e.g.}, play music) as well as pertinent entities needed to further refine the task (\textit{e.g.}, which playlist?).}

Traditional commercial voice assistants conventionally handle user utterances by conducting \textit{intent} detection and \textit{slot} extraction tasks separately. For example, given the utterance \textit{Play Paradise by Coldplay}, the semantic parsing module processes it in two stages: \textit{a)} initially determining the user's intent as \texttt{IN:PLAY\_MUSIC}, and then \textit{b)} recognizing task-specific named entities \textit{Paradise} and \textit{Coldplay}, respectively tagging these elements (\textit{slots}) as \texttt{SL:MUSIC\_TRACK\_TITLE} and \texttt{SL:MUSIC\_ARTIST\_NAME}. Intent detection has traditionally been approached as text classification, where the entire utterance serves as input, while slot recognition has been formulated as a sequence tagging challenge \citep{Mesnil2015,Liu2016AttentionBasedRN,goo-etal-2018-slot}.

Annotations generated by these traditional semantic parsers 
only support a single intent per utterance and a list of non-overlapping slots exclusively composed by tokens from the input. While this flat semantic representation suffices for handling straightforward utterances, it falls short in adequately representing user queries that involve \textit{compositional} requests. For instance, the query \textit{How long will it take to drive from my apartment to San Diego?} necessitates first identifying \textit{my apartment} (\texttt{IN:GET\_LOCATION\_HOME}) before estimating the duration to the destination (\texttt{IN:GET\_ESTIMATED\_DURATION}). Hence, there is a requirement for a semantic representation capable of managing multiple intents per utterance, where slots encapsulate nested intents. 

\begin{figure}
\includegraphics[width=\columnwidth]{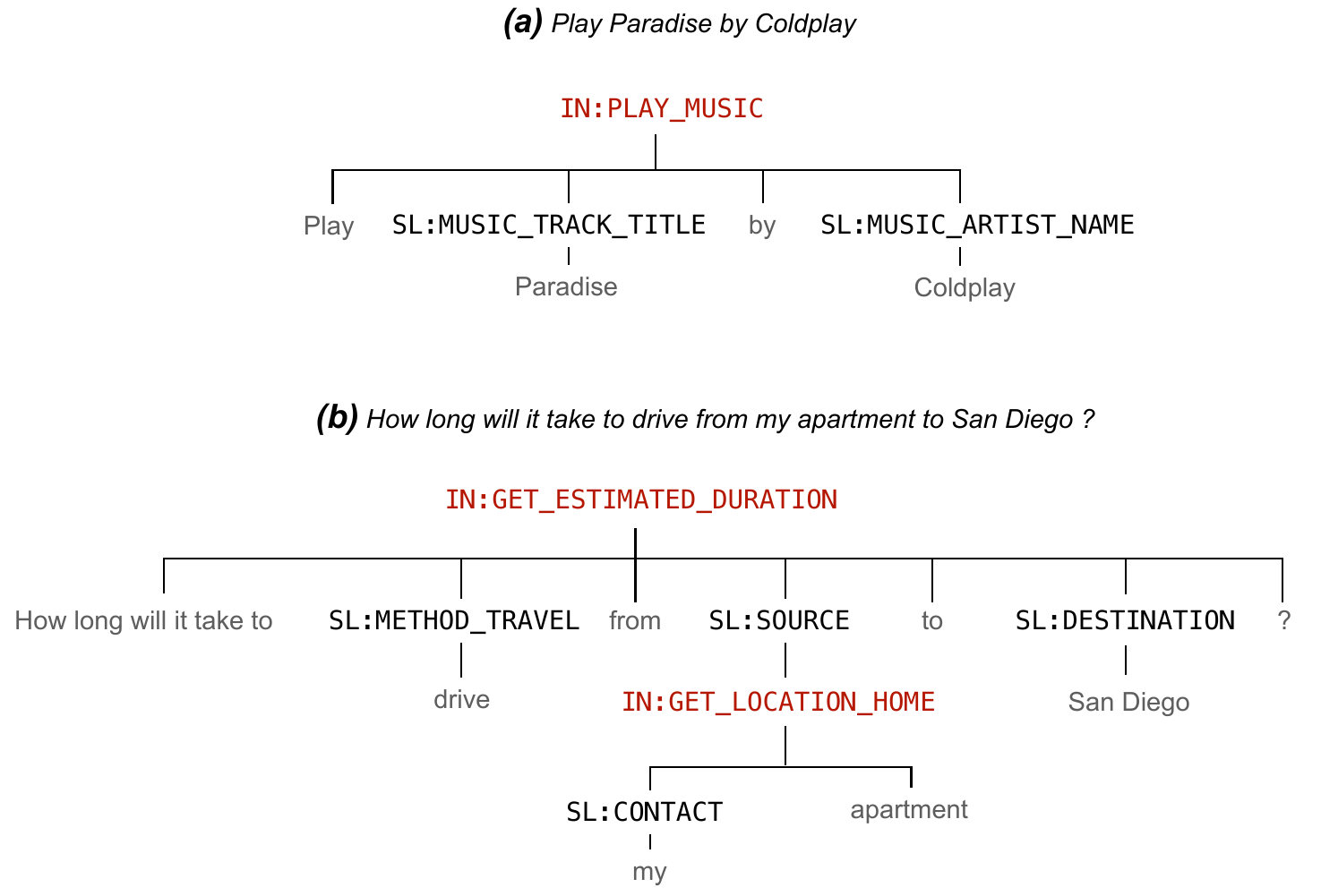}
\caption{Flat and compositional TOP annotations of utterances from music and navigation domains, respectively. Note that intents and slots are respectively prefixed with \texttt{IN:} and \texttt{SL:}.}
\label{fig:trees}
\end{figure}

In order to represent more complex utterances, \citet{gupta-etal-2018-semantic-parsing} introduced the \textit{task-oriented parsing} (TOP) formalism: a hierarchical annotation scheme expressive enough to describe the task-specific semantics of nested intents and model compositional queries. In Fig.~\ref{fig:trees}, we illustrate how the intents and slots of utterances mentioned in the previous examples can be represented using the TOP annotation.

An advantage of the TOP representation is its ease of annotation and parsing compared to more intricate semantic formalisms such as \textit{logical forms} \citep{Zelle96} or \textit{abstract meaning representations} (AMR) \citep{banarescu-etal-2013-abstract}. In fact, its similarity to a \textit{syntactic constituency} tree enables the adaptation of algorithms from the \textit{constituency parsing} literature to process task-oriented requests. This was the driving force behind \citet{gupta-etal-2018-semantic-parsing}'s initial proposal to modify the \textit{shift-reduce} constituency parser introduced by \citet{Dyer2016} for generating TOP annotations.

Alternatively, \citet{gupta-etal-2018-semantic-parsing} also proposed the application of different sequence-to-sequence models \citep{wiseman-rush-2016-sequence,pmlr-v70-gehring17a,Vaswani2017} for parsing compositional queries.  Sequence-to-sequence models comprise a specific neural architecture tasked with predicting a sequence of output tokens based on an input sequence of items. After conducting empirical comparisons between the shift-reduce technique and various sequence-to-sequence models for parsing compositional queries, they determined that the shift-reduce parser surpassed other methods and was the only approach capable of guaranteeing that the output representation adhered to a well-formed TOP tree. This superiority can be largely attributed to the fact that, unlike sequence-to-sequence models, shift-reduce algorithms adhere to grammar constraints throughout the parsing process and exhibit an inductive bias towards tree structures, resulting in enhanced performance.

Although sequence-to-sequence approaches may produce flawed representations, recent advancements \citep{rogali2020,aghajanyan-etal-2020-conversational} have substantially enhanced their performance by leveraging \textit{Transformer} neural networks \citep{Vaswani2017} in conjunction with pre-trained language models such as RoBERTa \citep{roberta} or BART \citep{lewis-etal-2020-bart}. Consequently, they have emerged as the most accurate approach to date for generating TOP tree structures.

This article presents further advancements in the realm of shift-reduce semantic parsing for natural language understanding. Specifically, we enhance the initial framework introduced by \citet{gupta-etal-2018-semantic-parsing}, which relied on the \textit{top-down} transition system \citep{Dyer2016} and a \textit{Stack-LSTM}-based neural architecture \citep{dyer-etal-2015-transition}. Firstly, we implement a more robust neural model based on \textit{Stack-Transformers} \citep{fernandez-astudillo-etal-2020-transition}, enabling the accurate modeling of shift-reduce systems within a Transformer-based neural architecture. Secondly, we adapt the \textit{bottom-up} and \textit{in-order} transition systems \citep{nonbinary,Liu2017} from the constituency parsing literature to task-oriented semantic parsing. Lastly, we empirically evaluate these alternatives, along with the top-down algorithm, on our neural architecture. Our findings demonstrate that the in-order transition system achieves the highest accuracy on the Facebook TOP benchmark \citep{gupta-etal-2018-semantic-parsing,chen-etal-2020-low}, even outperforming the most robust sequence-to-sequence baselines.

In summary, our contributions in this article are as follows:
\begin{itemize}
 \item  We develop innovative shift-reduce semantic parsers for task-oriented dialogues utilizing Stack-Transformers and \textit{deep contextualized word embeddings} derived from RoBERTa.
 \item  We adapt various transition systems from the constituency parsing literature to handle TOP annotations and conduct a comprehensive comparison against the original top-down approach, demonstrating the superiority of the in-order algorithm across all scenarios.
 \item  We evaluate our approach on both low-resource and high-resource settings of the Facebook TOP datasets, pushing the boundaries of the state of the art in task-oriented parsing and narrowing the divide with sequence-to-sequence models.
 \item  Our system’s source code is freely available at \url{https://github.com/danifg/ShiftReduce-TOP}. 
\end{itemize}

The remainder of this article is organized as follows:
In Section~\ref{sec:related}, we provide an overview of prior research on semantic parsing for task-oriented compositional queries.
Section~\ref{sec:shift-reduce} outlines our proposed approach, beginning with an exposition of the transition-based algorithms adapted from constituency parsing, followed by a detailed description of the Stack-Transformer-based neural model. Section~\ref{sec:experiments} presents the experiments conducted with the three transition systems using the proposed neural architecture as a testing platform, along with a comprehensive analysis of their performance. Finally, concluding remarks are presented in Section~\ref{sec:conclusion}.

\section{Related work}
\label{sec:related}

The hierarchical semantic representation introduced by \citet{gupta-etal-2018-semantic-parsing} to address compositional queries spurred the adaptation of parsing algorithms initially developed for constituency parsing, such as the Stack-LSTM-based shift-reduce parser \citep{Dyer2016}. Additionally, \citet{gupta-etal-2018-semantic-parsing} proposed sequence-to-sequence models for this task, including those based on \textit{convolutional neural networks} (CNNs) \citep{pmlr-v70-gehring17a}, \textit{long short-term memory} (LSTM) neural networks \citep{wiseman-rush-2016-sequence}, and Transformers \citep{Vaswani2017}. Although sequence-to-sequence methods were originally devised for machine translation \citep{Sutskever2014}, they were also adapted to constituency parsing by first linearizing the tree structure \citep{Vinyals2015}.

Given that the shift-reduce parser initially emerged as the leading method for generating TOP representations, \citet{Einolghozati2019ImprovingSP} opted to enhance the original system by incorporating an ensemble of seven parsers, contextualized word embeddings extracted from ELMo \citep{peters-etal-2018-deep}, and a language model ranker. Concurrently, \citet{pasupat-etal-2019-span} modified the span-based constituency parser proposed by \citet{stern-etal-2017-minimal} to process utterances into TOP trees, achieving promising results without the use of deep contextualized word embeddings.

\new{While sequence-to-sequence models initially lagged behind all available semantic parsing methods, recent advancements have substantially improved their performance in constructing TOP representations. Notably, \citet{rogali2020} devised a sequence-to-sequence architecture bolstered by a Pointer Generator Network \citep{see-etal-2017-get} and a RoBERTa-based encoder \citep{roberta}. This neural architecture emerged as the state of the art in task-oriented semantic parsing and has since been adopted and extended by subsequent studies. Among them, \citet{aghajanyan-etal-2020-conversational} and \citet{chen-etal-2020-low} proposed simplifying the target sequence by eliminating input tokens that are not slot values, while also initializing both the encoder and the decoder with the pre-trained sequence-to-sequence model BART \citep{lewis-etal-2020-bart}. Furthermore, non-autoregressive variants of the sequence-to-sequence architecture introduced by \citet{rogali2020} have been presented as well \citep{zhu-etal-2020-dont,babu-etal-2021-non,shrivastava-etal-2021-span-pointer,oh-etal-2022-improving}. Additionally, \citet{shrivastava-etal-2023-retrieve} recently enhanced sequence-to-sequence models with a scenario-based approach, where incomplete intent-slot templates are available in advance and can be retrieved after identifying the utterance's scenario. Meanwhile, \citet{wang-etal-2023-treepiece} chose to enhance the efficiency of sequence-to-sequence models by generating subtrees as output tokens at each decoding step.}

Diverging from the current mainstream trends, we push forward the boundaries of research in shift-reduce task-oriented parsing by crafting a novel approach grounded in a more accurate transition system and implemented on a more robust neural architecture. As a result, our system surpasses even the strongest sequence-to-sequence baselines.

\new{Simultaneously with our research, \citet{do-etal-2023-structsp} have developed a two-staged approach that demonstrates remarkable results. Initially, they enhance standard pre-trained language models through fine-tuning, incorporating additional hierarchical semantic information. Subsequently, the resulting model is integrated with a recursive insertion-based mechanism \citep{Mansimov2021}, constrained by grammar information. Specifically, grammar rules extracted from the training dataset are employed to prune unpromising predictions during the parsing process \citep{10.1007/978-3-031-35320-8_24}. It is worth noting that these contributions are orthogonal to our approach and could certainly enhance its performance.}

\section{Methodology}
\label{sec:shift-reduce}
This section outlines our proposed approach. Specifically, we elaborate on the transition-based algorithms adapted from the constituency parsing literature to handle task-oriented utterances, as well as the neural architecture serving as the foundation of our system.

\subsection{Transition systems for task-oriented semantic parsing}

In task-oriented semantic parsing, the objective is to transform an input utterance comprising $n$ words, denoted as $X=w_1, \dots, w_n$, into a semantic representation—in our case, a TOP tree $Y$. Similar to syntactic constituency representations, $Y$ is a rooted tree consisting of tokens $w_1, \dots w_n$ as its leaves and a collection of internal nodes (referred to as \textit{constituents}) hierarchically structured above them. These constituents are denoted as tuples $(N, W)$, where $W$ represents the set of tokens covered by its span, and $N$ denotes the \textit{non-terminal} label. For example, (\texttt{SL:SOURCE}, \{\textit{my}, \textit{apartment}\}) and (\texttt{SL:DESTINATION}, \{\textit{San}, \textit{Diego}\}) are constituents extracted from the TOP tree depicted in Fig.~\ref{fig:trees}(b). Additionally, in our specific scenario, two distinct types of constituents emerge: \textit{intents} and \textit{slots}, with non-terminal labels respectively prefixed with \texttt{IN:} and \texttt{SL:}. Finally, tree structures must adhere to certain constraints to be deemed a valid TOP representation:
\begin{itemize}
    \item The root constituent, which encompasses the entire utterance, must be an intent node.
\item Only tokens and/or slot constituents can serve as child nodes of an intent node.
\item A slot node may have either words (one or several) or a single intent constituent as child nodes.
\end{itemize}

To process the input utterance, we employ shift-reduce parsers, initially introduced for dependency and constituency parsing \citep{yamada03,sagae05}. These parsers construct the target tree incrementally by executing a sequence of actions that analyze the input utterance from left to right. Specifically, shift-reduce parsers are characterized by a non-deterministic \textit{transition system}, which defines the necessary data structures and the set of operations required to complete the parsing process; and an \textit{oracle}, which selects one of these actions deterministically at each stage of the parsing process. Formally, a transition system is represented as a quadruple $S = (C,c_0,C_f,T)$, where:
\begin{itemize}
\item $C$ denotes the set of possible \textit{state configurations}, defining the data structures necessary for the parser.
\item $c_0$ represents the \textit{initial configuration} of the parsing process.
\item $C_f$ is the set of \textit{final configurations} reached at the end of the parsing process.
\item $T$ signifies the set of available \textit{transitions} (or \textit{actions}) that can be applied to transition the parser from one state configuration to another.
\end{itemize}
\noindent Moreover, during training, a rule-based oracle $o$, given the gold parse tree $Y_g$, selects action $a_t$ for each state configuration $c_{t}$ at each time step $t$: $a_t=o(c_{t}, Y_g)$. Once the model is trained, it approximates the oracle during decoding.

We can utilize a transition system $S$ along with an oracle $o$ to parse the utterance $X$: commencing from the initial configuration $c_0$, a sequence of transitions ${a_0, \dots, a_{m-1}}$ (determined by the oracle at each time step $t$) guides the system through a series of state configurations ${c_0, \dots, c_m}$ until a final configuration is reached ($c_m \in C_f$). At this stage, the utterance will have been fully processed, and the parser will generate a valid TOP tree $Y$. Various transition systems exist in the literature on constituency parsing. In addition to the algorithm employed by \citet{gupta-etal-2018-semantic-parsing}, we have adapted two other transition systems for task-oriented semantic parsing, which we elaborate on in the subsequent sections.

\paragraph{Top-down transition system}
Initially conceived by \citet{Dyer2016} for constructing constituency trees in a top-to-bottom fashion, this transition system was later adapted by \citet{gupta-etal-2018-semantic-parsing} to accommodate TOP tree representations. The top-down transition system comprises the following components:
\begin{itemize}

     \item State configurations within $C$ are structured as {$c=\langle {\Sigma} , {B} \rangle$}, where $\Sigma$ denotes a \textit{stack} (responsible for storing non-terminal symbols, constituents, and partially processed tokens), and $B$ represents a \textit{buffer} (containing unprocessed tokens to be read from the input).
    \item At the initial configuration $c_0$, the buffer $B$ encompasses all tokens from the input utterance, while the stack $\Sigma$ remains empty.
     \item Final configurations within $C_f$ are structured as {$c=\langle [ I ], \emptyset \rangle$}, where the buffer is empty (indicating that all words have been processed), and the stack contains a single item $I$. This item represents an intent constituent spanning the entire utterance, as the root node of a valid TOP tree must be an intent.
    \item The set of available transitions $T$ consists of three actions: 
    \begin{itemize}
       \item The \textsc{Non-Terminal-}\textit{L} transition involves pushing a non-terminal node labeled \textit{L} onto the stack, transitioning the system from state configurations of the form $\langle {\Sigma}, B \rangle$ to $\langle {\Sigma | L }, B  \rangle$ (where $\Sigma | L$ denotes a stack with item $L$ placed on top and $\Sigma$ as the tail). Unlike in constituency parsing, this transition can generate intent and slot non-terminals (with labels \textit{L} prefixed with \texttt{IN:} or \texttt{SL:}, respectively). Therefore, it must adhere to specific constraints to produce a well-formed TOP tree:

            \begin{itemize}
                \item Since the root node must be an intent constituent, the first \textsc{Non-Terminal-}\textit{L} transition must introduce an intent non-terminal onto the stack.
                \item A \textsc{Non-Terminal-}\textit{L} transition that inserts an intent non-terminal onto the stack is permissible only if the last pushed non-terminal was a slot, performed in the preceding state configuration. This condition ensures that the resulting intent constituent from this transition becomes the sole child node of that preceding slot, as required by the TOP formalism.
                \item A \textsc{Non-Terminal-}\textit{L} transition adding a slot non-terminal to the stack is allowed only if the last inserted non-terminal was an intent.
            \end{itemize}

     \item  A \textsc{Shift} action is employed to retrieve tokens from the input by transferring words from the buffer to the stack. This operation transitions the parser from state configurations $\langle {\Sigma}, {w_i | B}  \rangle $ to $\langle {\Sigma | w_i} , {B}  \rangle$ (where ${w_i | B}$ denotes a buffer with token $w_i$ on top and $B$ as the tail, and conversely, $\Sigma | w_i$ represents a stack with $\Sigma$ as the tail and $w_i$ as the top). This transition is permissible only if the buffer is not empty. Specifically for task-oriented semantic parsing, this action will not be available in state configurations where the last non-terminal added to the stack was a slot, and an intent constituent was already created as its first child node. This constraint ensures that slot constituents have only one intent as their child node.

    \item Additionally, a \textsc{Reduce} transition is necessary to construct a new constituent by removing all items (including tokens and constituents) from the stack until a non-terminal symbol is encountered, then grouping them as child nodes of that non-terminal. This results in a new constituent placed on top of the stack, transitioning the parser from configurations $\langle {\Sigma} | L| e_k| \dots | e_0 , B \rangle$ to $\langle {\Sigma} | L_{e_k \dots  e_0} , B \rangle$ (where $L_{e_k \dots  e_0}$ denotes a constituent with non-terminal label $L$ and child nodes $e_k \dots  e_0$). This transition can be executed only if there are at least one non-terminal symbol and one item (token or constituent) in the stack.
        \end{itemize}
\noindent Please note that the original work by \citet{gupta-etal-2018-semantic-parsing} did not provide specific transition constraints tailored to generating valid TOP representations. Therefore, we undertook a complete redesign of the original top-down algorithm \citep{Dyer2016} for task-oriented semantic parsing to incorporate these task-specific transition constraints.
\end{itemize}
Finally, Table~\ref{fig:example} illustrates how the top-down algorithm parses the utterance depicted in Fig.~\ref{fig:trees}(a). It demonstrates the step-by-step construction of each constituent, which involves defining the non-terminal label, reading and/or creating all corresponding child nodes, and then reducing all items within its span.

\begin{table}
\footnotesize
\begin{tabular}{@{\hskip 0pt}l@{\hskip 0pt}c@{\hskip 0pt}c@{\hskip 0pt}}
\toprule
Transition & Stack & Buffer \\
\midrule
\vspace*{3pt}
 & [ ] & [Play, Paradise, by, Coldplay]   \\
 \vspace*{3pt}
\textsc{NT-}\texttt{\tiny IN:PLAY\_MUSIC} & [ \texttt{IN:PLAY\_MUSIC} ] & [Play, Paradise, by, Coldplay]   \\ 
\vspace*{3pt}
\textsc{Shift} & [ \texttt{IN:PLAY\_MUSIC}, Play ] & [ Paradise, by, Coldplay ]   \\
\vspace*{3pt}
\textsc{NT-}\texttt{\tiny SL:TITLE} & [ \texttt{IN:PLAY\_MUSIC}, Play, \texttt{SL:TITLE} ] & [ Paradise, by, Coldplay ]   \\
\vspace*{3pt}
\textsc{Shift} & [ \texttt{IN:PLAY\_MUSIC}, Play, \texttt{SL:TITLE}, Paradise ] & [ by, Coldplay ]   \\
\vspace*{3pt}
\textsc{Reduce} & [ \texttt{IN:PLAY\_MUSIC}, Play, \texttt{SL:TITLE}$_{Paradise}$ ] & [ by, Coldplay ]   \\
\vspace*{3pt}
\textsc{Shift} & [ \texttt{IN:PLAY\_MUSIC}, Play, \texttt{SL:TITLE}$_{Paradise}$, by ] & [ Coldplay ]   \\
\vspace*{3pt}
\textsc{NT-}\texttt{\tiny SL:ARTIST} & [ \texttt{IN:PLAY\_MUSIC}, Play, \texttt{SL:TITLE}$_{Paradise}$, by, \texttt{SL:ARTIST} ] & [ Coldplay ]   \\
\vspace*{3pt}
\textsc{Shift} & [\texttt{IN:PLAY\_MUSIC}, Play, \texttt{SL:TITLE}$_{Paradise}$, by, \texttt{SL:ARTIST}, Coldplay] & [ ]   \\
\vspace*{3pt}
\textsc{Reduce} & [ \texttt{IN:PLAY\_MUSIC}, Play, \texttt{SL:TITLE}$_{Paradise}$, by, \texttt{SL:ARTIST}$_{Coldplay}$ ] & [ ]   \\
\vspace*{3pt}
\textsc{Reduce} & [ \texttt{IN:PLAY\_MUSIC}$_{Play\ \texttt{SL:TITLE}\ by\ \texttt{SL:ARTIST}}$ ] & [ ]   \\
\bottomrule
\end{tabular}
\caption{Top-down transition sequence and state configurations (represented by the stack and the buffer)
for 
producing the
TOP tree in Fig.~\ref{fig:trees}(a).
\textsc{NT-}\textit{\tiny L} stands for \textsc{Non-Terminal-}\textit{L} and slot labels have been abbreviated from  \texttt{SL:MUSIC\_TRACK\_TITLE} and \texttt{SL:MUSIC\_ARTIST\_NAME} to \texttt{SL:TITLE} and \texttt{SL:ARTIST}, respectively.} \label{fig:example}   
\end{table}

\paragraph{Bottom-up transition system} 
In contrast to the top-down approach, shift-reduce algorithms traditionally perform constituency parsing by building trees from bottom to top. Therefore, we have also adapted the bottom-up transition system developed by \citet{nonbinary} for task-oriented semantic parsing. Unlike classic bottom-up constituency parsing algorithms \citep{sagae05,zhu-etal-2013-fast}, this transition system does not require prior binarization of the gold tree during training or subsequent recovery of the non-binary structure after decoding. Specifically, the non-binary bottom-up transition system comprises:
\begin{itemize}
    \item State configurations have the form {$c=\langle {\Sigma} , {B}, f \rangle$}, where $\Sigma$ is a \textit{stack}, $B$ is a \textit{buffer}, as described for the top-down algorithm, and $f$ is a boolean variable indicating whether a state configuration is terminal or not.
\item In the initial configuration $c_0$, the buffer contains the entire user utterance, the stack is empty, and $f$ is \textit{false}.
\item Final configurations in $C_f$ have the form {$c=\langle [I], \emptyset, \textit{true} \rangle$}, where the stack holds a single intent constituent, the buffer is empty, and $f$ is \textit{true}. Following a bottom-up algorithm, we can continue building constituents on top of a single intent node in the stack, even when it spans the whole input utterance. To avoid that, this transition system requires the inclusion of variable $f$ in state configurations to indicate the end of the parsing process.
\item Actions provided by this bottom-up algorithm are as follows: 
\begin{itemize}
  \item  Similar to the top-down approach, a \textsc{Shift} action moves tokens from the buffer to the stack, transitioning the parser from state configurations $\langle {\Sigma}, {w_i | B}, \textit{false}  \rangle $ to $\langle {\Sigma | w_i} , {B}, \textit{false}  \rangle$. This operation is not permissible under the following conditions:
        \begin{itemize}
          \item When the buffer is empty and there are no more words to read.
          \item When the top item on the stack is an intent node and, since slots must have only one intent child node, the parser needs to build a slot constituent on top of it before shifting more input tokens.
          \end{itemize}
    \item A \textsc{Reduce\#k-}\textit{L} transition (parameterized with the non-terminal label \textit{L} and an integer \textit{k}) is used to create a new constituent by popping \textit{k} items from the stack and combining them into a new constituent on top of the stack. This transitions the parser from state configurations $\langle {\Sigma} | e_{k-1}| \dots | e_0 , B, \textit{false} \rangle$ to $\langle {\Sigma} | L_{e_{k-1} \dots  e_0} , B, \textit{false} \rangle$. To ensure a valid TOP representation, this transition can only be applied under the following conditions:
    \begin{itemize}
  \item  When the \textsc{Reduce\#k-}\textit{L} action creates an intent constituent (\textit{i.e.}, \textit{L} is prefixed with \texttt{IN:}), it is permissible only if there are no intent nodes among the $k$ items popped from the stack (as an intent constituent cannot have other intents as child nodes).
  \item  When the \textsc{Reduce\#k-}\textit{L} transition builds a slot node (\textit{i.e.}, \textit{L} is prefixed with \texttt{SL:}), it is allowed only if there are no slot constituents among the $k$ elements affected by this operation (as slots cannot have other slots as child nodes). Additionally, if the item on top of the stack is an intent node, only the \textsc{Reduce} action with $k$ equal to 1 is permissible (since slots can only contain a single intent constituent as a child node).
  \end{itemize}
  
    \item Lastly, a \textsc{Finish} action is used to signal the end of the parsing process by changing the value of $f$, transitioning the system from configurations $\langle {\Sigma}, B, \textit{false} \rangle$ to final configurations $\langle {\Sigma}, B, \textit{true} \rangle$. This operation is only allowed if the stack contains a single intent constituent and the buffer is empty.

\end{itemize}
\end{itemize}
Finally, Table~\ref{fig:example2} illustrates how this shift-reduce parser processes the utterance in Fig.~\ref{fig:trees}(a), constructing each constituent from bottom to top by assigning the non-terminal label after all child nodes are fully assembled in the stack.

\begin{table}
\begin{tabular}{@{\hskip 0pt}l@{\hskip 2pt}c@{\hskip 0pt}c@{\hskip 2pt}c@{\hskip 0.1pt}}
\toprule
Transition & Stack & Buffer & $f$\\
\midrule
\vspace*{3pt}
 & [ ] & [Play, Paradise, by, Coldplay]   & \textit{false}\\
\vspace*{3pt}
\textsc{Shift} & [ Play ] & [Paradise, by, Coldplay] & \textit{false}   \\
\vspace*{3pt}
\textsc{Shift} & [ Play, Paradise ] & [ by, Coldplay ]  & \textit{false} \\
\vspace*{3pt}
\textsc{Re\#1-}\texttt{\tiny SL:TITLE} & [ Play, \texttt{SL:TITLE}$_{Paradise}$ ] & [ by, Coldplay ] & \textit{false}  \\
\vspace*{3pt}
\textsc{Shift} & [ Play, \texttt{SL:TITLE}$_{Paradise}$, by ] & [ Coldplay ]  & \textit{false} \\
\vspace*{3pt}
\textsc{Shift} & [ Play, \texttt{SL:TITLE}$_{Paradise}$, by, Coldplay ] & [ ]  & \textit{false} \\
\vspace*{3pt}
\textsc{Re\#1-}\texttt{\tiny SL:ARTIST} & [Play, \texttt{SL:TITLE}$_{Paradise}$, by, \texttt{SL:ARTIST}$_{Coldplay}$] & [ ]  & \textit{false} \\
\vspace*{3pt}
\textsc{Re\#4-}\texttt{\tiny IN:PLAY\_MUSIC} & [ \texttt{IN:PLAY\_MUSIC}$_{Play\ \texttt{SL:TITLE}\ by\ \texttt{SL:ARTIST}}$ ] & [ ] & \textit{false}  \\
\vspace*{3pt}
\textsc{Finish} & [ \texttt{IN:PLAY\_MUSIC}$_{Play\ \texttt{SL:TITLE}\ by\ \texttt{SL:ARTIST}}$ ] & [ ]  & \textit{true} \\
\bottomrule
\end{tabular}
\caption{Transition sequence and state configurations (represented by the stack, buffer and variable $f$)
for 
building the
TOP semantic representation in Fig.~\ref{fig:trees}(a) following a non-binary bottom-up approach. \textsc{Re\#k-}\textit{\tiny L} stands for \textsc{Reduce\#k-}\textit{L}.
} \label{fig:example2}   

\end{table}

\paragraph{In-order transition system}
\label{sec:inorder}
Alternatively to the top-down and bottom-up strategies, \citet{Liu2017} introduced the \textit{in-order} transition system for constituency parsing. We have tailored this algorithm for parsing task-oriented utterances. Specifically, the proposed in-order transition system consists of:
\begin{itemize}
    \item Configurations maintain the same format as the bottom-up algorithm (\textit{i.e.,} {$c=\langle {\Sigma} , {B}, f \rangle$}).
\item In the initial configuration $c_0$, the buffer contains the entire user utterance, the stack is empty, and the value of $f$ is \textit{false}.
\item Final configurations take the form {$c=\langle [I], \emptyset, \textit{true} \rangle$}. Similar to the bottom-up approach, the in-order algorithm may continue creating additional constituents above the intent node left on the stack indefinitely. Hence, a flag is necessary to indicate the completion of the parsing process.
\item The available transitions are adopted from both top-down and bottom-up algorithms, but some of them exhibit different behaviors or are applied in a different order:
    
\begin{itemize}
\item A \textsc{Non-Terminal-}\textit{L} transition involves pushing a non-terminal symbol \textit{L} onto the stack, transitioning the system from state configurations represented as $\langle {\Sigma}, B, \textit{false} \rangle$ to $\langle {\Sigma | L }, B , \textit{false} \rangle$. However, unlike the top-down algorithm, this transition can only occur if the initial child node of the upcoming constituent is fully constructed on top of the stack. Furthermore, it must meet other task-specific constraints to generate valid TOP representations:

 \begin{itemize}
    \item A \textsc{Non-Terminal-}\textit{L} transition that introduces an intent non-terminal to the stack (\textit{i.e.}, \textit{L} prefixed with \texttt{IN:}) is valid only if its first child node atop the stack is not an intent constituent.
\item  A \textsc{Non-Terminal-}\textit{L} transition that places a slot non-terminal on the stack (\textit{i.e.}, \textit{L} prefixed with \texttt{SL:}) is permissible only if the fully-created item atop the stack is not a slot node.
    \end{itemize}
\item  Similarly to other transition systems, a \textsc{Shift} operation is used to retrieve tokens from the buffer. However, unlike those algorithms, this action is restricted if the upcoming constituent has already been labeled as a slot (by a non-terminal previously added to the stack) and its first child node is an intent constituent already present in the stack. This condition aims to prevent slot constituents from having more than one child node when the item at the top of the stack is an intent.

\item A \textsc{Reduce} transition is employed to generate intent or slot constituents. Specifically, it removes all elements from the stack until a non-terminal symbol is encountered, which is simultaneously replaced by the preceding item to form a new constituent at the top of the stack. Consequently, it guides the parser from state configurations represented as $\langle {\Sigma} | e_k| L| e_{k-1}| \dots | e_0 , B, \textit{false} \rangle$ to $\langle {\Sigma} | L_{e_k \dots  e_0} , B, \textit{false} \rangle$. This transition is only applicable if there is a non-terminal in the stack (preceded by its first child constituent according to the in-order algorithm). Additionally, this transition must comply with specific constraints for task-oriented semantic parsing:
\begin{itemize}
\item When the \textsc{Reduce} operation results in an intent constituent (as determined by the last non-terminal label added to the stack), it is permissible only if there are no intent nodes among the preceding $k-1$ items (since the first child $e_k$ already adheres to the TOP formalism, as verified during the application of the \textsc{Non-Terminal-}\textit{L} transition).

\item When the \textsc{Reduce} transition produces a slot constituent, it is allowed only if there are no other slot nodes within the preceding $k-1$ elements that will be removed by this operation. This condition also encompasses scenarios where the initial child node $e_k$ of the upcoming slot constituent is an intent and, since the \textsc{Shift} transition is not permitted under such circumstances, only the \textsc{Reduce} action can construct a slot with a single intent.

\end{itemize}

    \item Lastly, akin to the bottom-up approach, a \textsc{Finish} action is utilized to finalize the parsing process. This action is only permissible if the stack contains a single intent constituent and the buffer is empty.
 \end{itemize}
\end{itemize}

In Table~\ref{fig:example3}, we illustrate how the in-order strategy parses the user utterance depicted in Fig.~\ref{fig:trees}(a). While the top-down and bottom-up approaches can be respectively regarded as a \textit{pre-order} and \textit{post-order} traversal over the tree, this transition system constructs the constituency structure following an \textit{in-order} traversal, addressing the drawbacks of the other two alternatives. The in-order strategy creates each constituent by determining the non-terminal label after its first child is completed in the stack, and then processing the remaining child nodes. Unlike the top-down approach, which assigns the non-terminal label before reading the tokens composing its span, the in-order algorithm can utilize information from the first child node to make a better choice regarding the non-terminal label. On the other hand, the non-binary bottom-up strategy must simultaneously determine the non-terminal symbol and the left span boundary of the future constituent once all child nodes are completed in the stack. Despite having local information about already-built subtrees, the bottom-up strategy lacks global guidance from top-down parsing, which is essential for selecting the correct non-terminal label. Additionally, determining span boundaries can be challenging when the target constituent has a long span, as \textsc{Reduce\#k-}\textit{L} transitions with a high \textit{k} value are less frequent in the training data and thus harder to learn. The in-order approach avoids these drawbacks by predicting the non-terminal label and marking the left span boundary after creating its first child. In Section~\ref{sec:experiments}, we will empirically demonstrate that, in practice, the advantages of the in-order transition system result in substancial accuracy improvements compared to the other two alternatives.

\begin{table}

\footnotesize
\vspace*{13pt}
\begin{tabular}{@{\hskip 0pt}l@{\hskip 0pt}c@{\hskip 0pt}c@{\hskip 5pt}c@{\hskip 1pt}}
\toprule
Transition & Stack & Buffer & $f$\\
\midrule
\vspace*{3pt}
 & [ ] & [ Play, Paradise, by, Coldplay ]   & \textit{false}\\
\vspace*{3pt}
\textsc{Shift} & [ Play ] & [ Paradise, by, Coldplay ] & \textit{false}   \\
\vspace*{3pt}
\textsc{NT-}\texttt{\tiny{IN:PLAY\_MUSIC}} & [ Play, \texttt{IN:PLAY\_MUSIC} ] & [ Paradise, by, Coldplay ] & \textit{false}   \\
\vspace*{3pt}
\textsc{Shift} & [ Play, \texttt{IN:PLAY\_MUSIC}, Paradise ] & [ by, Coldplay ] & \textit{false}   \\
\vspace*{3pt}
\textsc{NT-}\texttt{\tiny{SL:TITLE}} & [ Play, \texttt{IN:PLAY\_MUSIC}, Paradise, \texttt{SL:TITLE} ] & [ by, Coldplay ] & \textit{false}   \\
\vspace*{3pt}
\textsc{Reduce} & [ Play, \texttt{IN:PLAY\_MUSIC}, \texttt{SL:TITLE}$_{Paradise}$ ] & [ by, Coldplay ] & \textit{false}   \\
\vspace*{3pt}
\textsc{Shift} & [ Play, \texttt{IN:PLAY\_MUSIC}, \texttt{SL:TITLE}$_{Paradise}$, by ] & [ Coldplay ] & \textit{false}   \\
\vspace*{3pt}
\textsc{Shift} & [ Play, \texttt{IN:PLAY\_MUSIC}, \texttt{SL:TITLE}$_{Paradise}$, by, Coldplay ] & [ ] & \textit{false}   \\
\vspace*{3pt}
\textsc{NT-}\texttt{\tiny{SL:ARTIST}} & [ Play, \texttt{IN:PLAY\_MUSIC}, \dots , Coldplay, \texttt{SL:ARTIST} ] & [ ] & \textit{false}   \\
\vspace*{3pt}
\textsc{Reduce} & [ Play, \texttt{IN:PLAY\_MUSIC}, \dots , by, \texttt{SL:ARTIST}$_{Coldplay}$ ] & [ ]  & \textit{false} \\
\vspace*{3pt}
\textsc{Reduce} & [ \texttt{IN:PLAY\_MUSIC}$_{Play\ \texttt{SL:TITLE}\ by\ \texttt{SL:ARTIST}}$ ] & [ ]  & \textit{false} \\
\vspace*{3pt}
\textsc{Finish} & [ \texttt{IN:PLAY\_MUSIC}$_{Play\ \texttt{SL:TITLE}\ by\ \texttt{SL:ARTIST}}$ ] & [ ]  & \textit{true} \\
\bottomrule
\end{tabular}
\caption{In-order transition sequence and state configurations for
generating the
TOP representation in Fig.~\ref{fig:trees}(a). \textsc{NT-}\textit{\tiny L} stands for \textsc{Non-Terminal-}\textit{L} and slot labels have been respectively abbreviated from  \texttt{SL:MUSIC\_TRACK\_TITLE} and \texttt{SL:MUSIC\_ARTIST\_NAME} to \texttt{SL:TITLE} and \texttt{SL:ARTIST}.
} \label{fig:example3}   
\vspace*{13pt}

\end{table}

\subsection{Neural parsing model}
\label{sec:transformer}
Earlier shift-reduce systems in dependency parsing \citep{dyer-etal-2015-transition}, constituency parsing \citep{Dyer2016}, AMR parsing \citep{ballesteros-al-onaizan-2017-amr} and task-oriented semantic parsing \citep{gupta-etal-2018-semantic-parsing,Einolghozati2019ImprovingSP}  relied on \textit{Stack-LSTMs} for modeling state configurations.
These architectures are grounded in LSTM recurrent neural networks \citep{Hochreiter97},  which dominated the natural language processing community until \citet{Vaswani2017} introduced \textit{Transformers}. This neural architecture  offers a cutting-edge attention mechanism \citep{Bahdanau15} that outperforms LSTM-based systems and, unlike recurrent neural networks, can be easily parallelized. 
This motivated \citet{fernandez-astudillo-etal-2020-transition} to design \textit{Stack-Transformers}. In particular,
they use Stack-Transformers to replace Stack-LSTMs in shift-reduce dependency and AMR parsing, achieving remarkable gains in accuracy.

In our research, we leverage Stack-Transformers to represent the buffer and stack structures of the described transition systems, employing them to construct innovative shift-reduce task-oriented parsers. Specifically, we implement the following encoder-decoder architecture:

\begin{figure*}
\centering
\includegraphics[width=0.6\textwidth]{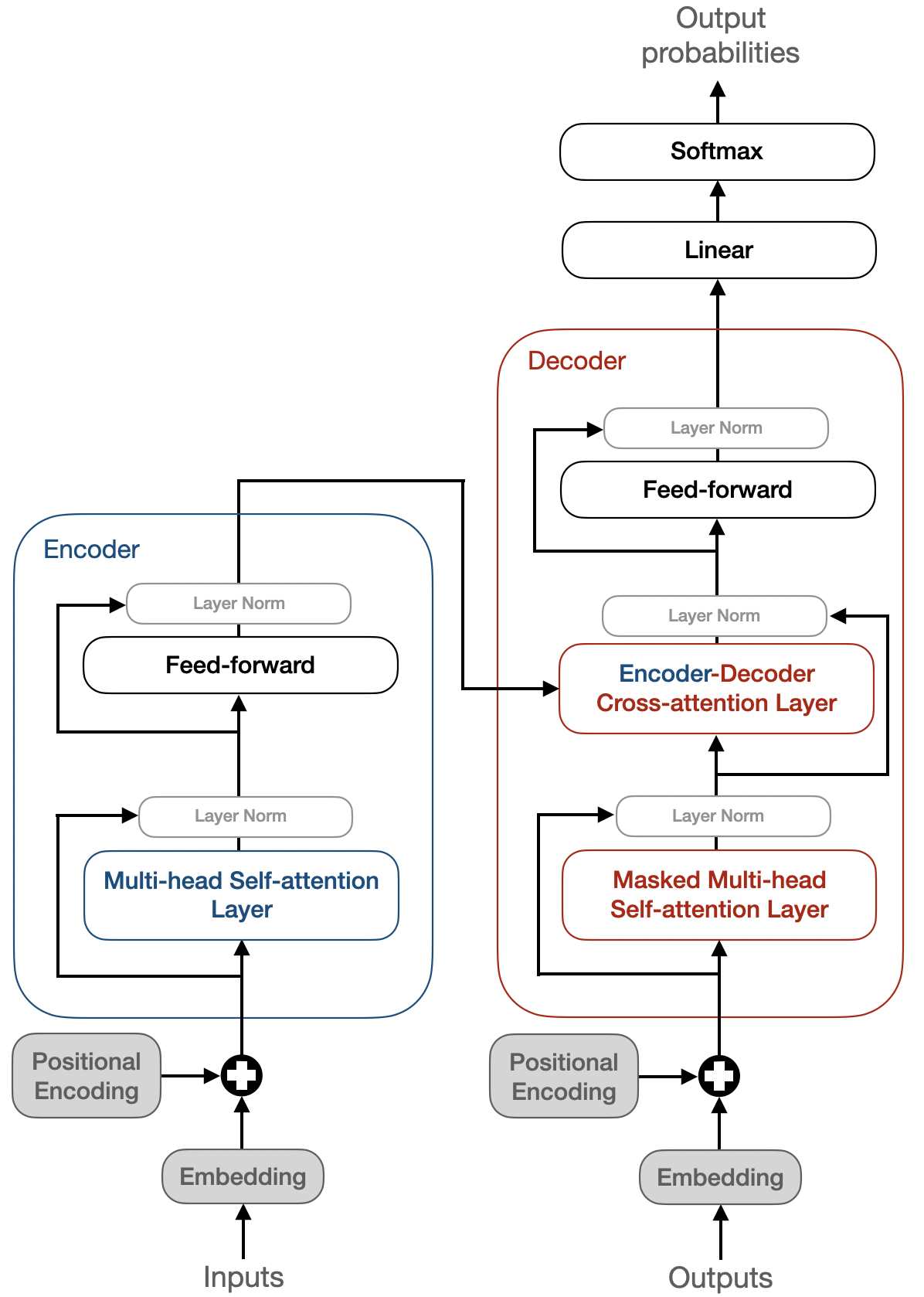}
\caption{\new{Transformer neural architecture introduced by  \citet{Vaswani2017}. Note that this neural network requires  the incorporation of \textit{positional encoding} for each input token to maintain sequential order, and \textit{Layer Norm} refers to the layer normalization technique proposed by \citet{layernormalization}.}}
\label{fig:network}
\end{figure*}

\paragraph{Encoder} 
Top-performing sequence-to-sequence approaches \citep{rogali2020,zhu-etal-2020-dont} directly use pre-trained models like RoBERTa  \citep{roberta} as the encoder in their neural architectures, conducting a task-specific fine-tuning during training. RoBERTa, short for "Robustly optimized BERT pretraining approach," employs the same Transformer architecture as BERT \citep{devlin-etal-2019-bert} and was pre-trained on masked word prediction using a large dataset.

Unlike strong sequence-to-sequence techniques, we adopt a less resource-consuming and greener strategy: 
we extract fixed weights from the pre-trained language model RoBERTa$_\textsc{Large}$\footnote{\url{https://huggingface.co/roberta-large}} 
 to initialize word embeddings, which remain frozen throughout the training process. Specifically, we use mean pooling (i.e., averaging the weights from wordpieces) to generate a word representation $e_i$ for each token $w_i$ in the input utterance $X=w_1, \dots, w_n$, resulting in the sequence $E=e_1, \dots, e_n$.

Next, we define the \textit{encoder} using a 6-layer Transformer with a hidden size of 256. Transformers utilize a \textit{multi-head self-attention layer} with multiple attention heads (four in our case) to assess the relevance of each input token relative to the other words in the utterance. The output of this layer is fed into a feed-forward layer, ultimately producing an encoder hidden state $h_i$ for each input word (represented as $e_i$). Therefore, given the sequence of word representations $E$, the encoding process yields the sequence of \textit{encoder hidden states} $H = h_1, \dots, h_n$. Fig.~\ref{fig:network} illustrates the Transformer neural architecture.

\paragraph{Decoder with Stack-Transformers} 
The decoder is responsible for generating the sequence of target actions $A = a_0, \dots, a_{m-1}$ to parse the input utterance $X$ according to a specific transition system $S$.

We use Stack-Transformers (with 6 layers, a hidden size of 256, and 4 attention heads) to effectively model the stack and buffer structures at each state configuration of the shift-reduce parsing process. In the original Transformer decoder model, a \textit{cross-attention layer} employs multiple \textit{attention heads} to attend to all input tokens and compute their compatibility with the last \textit{decoder hidden state} $q_t$ (which encodes the transition history). However, Stack-Transformers specialize one attention head to focus exclusively on tokens in the stack at state configuration $c_t$ and another head solely on the remaining words in the buffer at $c_t$. This specialization allows the Transformer to represent stack and buffer structures. 

\begin{figure}
\begin{center}
\includegraphics[width=\columnwidth]{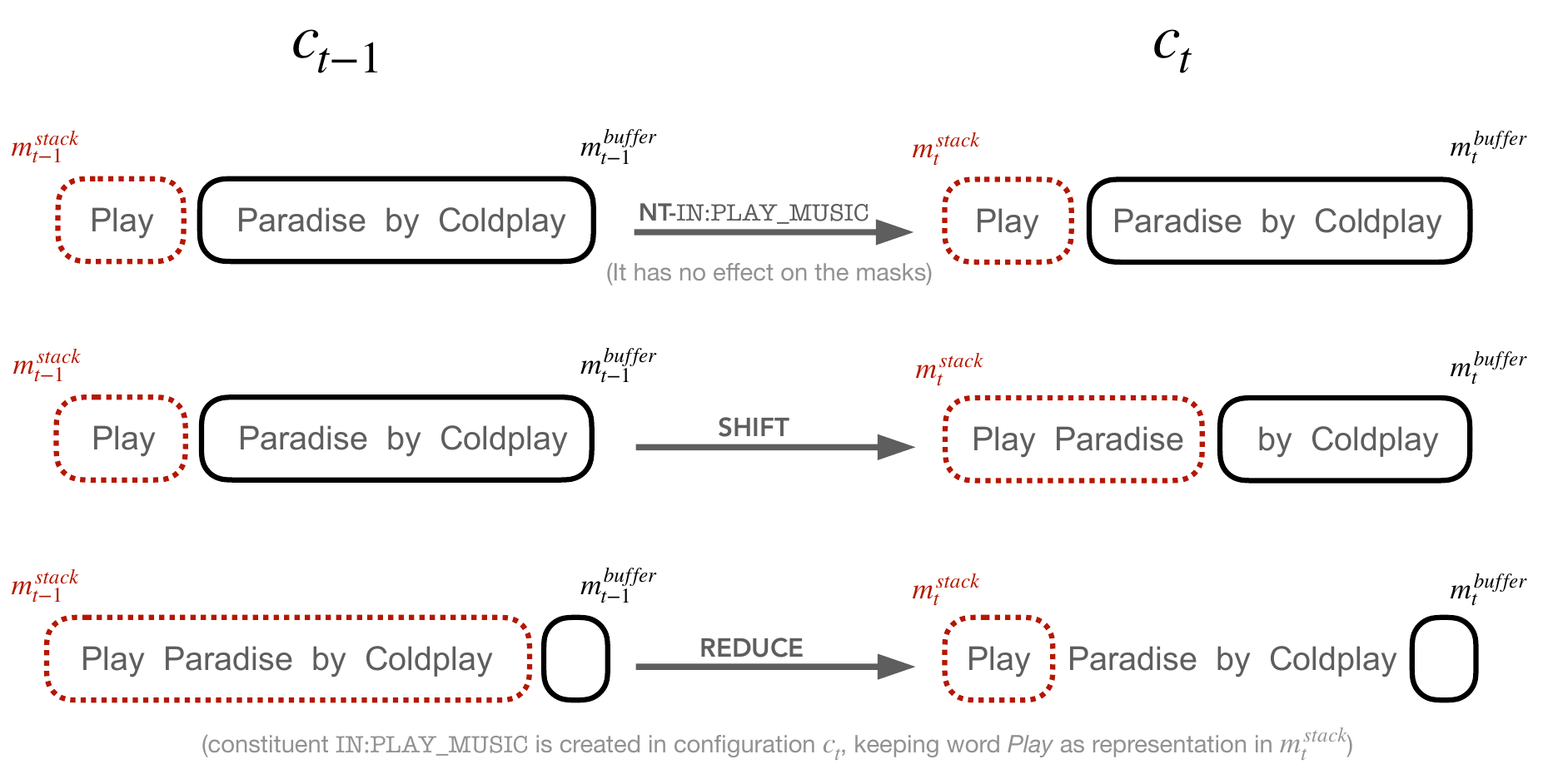}
\end{center}
\caption{Updates to the masks $m^{\textit{stack}}_t$ and $m^{\textit{buffer}}_t$ that reflect the effects of certain in-order transitions on the stack and buffer during the shift-reduce parsing process illustrated in Table~\ref{fig:example3}.}
\label{fig:masks}
\end{figure}

In practice, these dedicated stack and buffer attention heads are implemented using masks $m^{\textit{stack}}$ and $m^{\textit{buffer}}$ over the input. After applying the transition $a_{t-1}$ to state configuration $c_{t-1}$, these masks must be updated at time step $t$ to accurately represent the stack and buffer contents in the current state configuration $c_t$. To achieve this, we define how these masks are specifically modified for each transition system described in Section~\ref{sec:shift-reduce}:

\begin{itemize}
    \item If the action $a_{t-1}$ is a \textsc{Shift} transition, the first token in $m^{\textit{buffer}}$ will be masked out and added to $m^{\textit{stack}}$. This applies to all proposed transition systems, as the \textsc{Shift} transition behaves consistently across them. 
    \item When a \textsc{Non-terminal-}\textit{L} transition is applied, it affects the stack structure in $c_t$ but has no effect on $m^{\textit{stack}}$. This is because attention heads only attend to input tokens, and non-terminals are artificial symbols not present in the user utterance.
    \item For a \textsc{Reduce} transition (including the \textsc{Reduce\#k-}\textit{L} action from the non-binary bottom-up transition system), all tokens in $m^{\textit{stack}}$ that form the upcoming constituent will be masked out, except for the initial word representing the resultant constituent (since artificial non-terminals cannot be considered by the attention heads).
    
\end{itemize}
In Fig.~\ref{fig:masks}, we illustrate how these masks represent the content of the buffer and stack structures and how they are adjusted as the parser transitions from state configurations $c_{t-1}$ to $c_t$.
 
After encoding the stack and buffer in state configuration $c_t$ into masks $m^{\textit{stack}}_t$ and $m^{\textit{buffer}}_t$ (both represented as vectors with values of $-\infty$ or 0), the attention head $z^{\textit{stack}}_t$ (focused exclusively on the stack) is computed as follows:

\begin{equation}
    z^{stack}_t = \sum_{i=1}^{n}\alpha_{ti}(h_iW_d^V), \ \ 
 \alpha_{ti} = \frac{exp(\beta_{ti})}{\sum_{k=1}^{n} exp(\beta_{tk})}, \ \ \beta_{ti} = \frac{(q_{t}W_d^Q)(h_i{W_d^K}) ^T}{\sqrt{d}} + m^{\textit{stack}}_{ti}
\end{equation}
\noindent where $W_d^K$, $W_d^Q$ and $W_d^V$ are parameter matrices unique to each attention head, $d$ is the dimension of the resulting attention vector $z^{stack}_t$, and $\beta_{ti}$ is a compatibility function that measures the interaction between the decoder hidden state $q_{t}$ 
and each input token $w_i$ (represented by $h_i$).
By introducing the mask $m^{\textit{stack}}_t$ into the original equation to compute $\beta_{ti}$, this scoring function will only affect the words that are in the stack at time step $t$.

Similarly, the attention vector $z^{\textit{buffer}}_t$ (which only affects input tokens in the buffer in $c_t$) is calculated as:
\begin{equation}
    z^{\textit{buffer}}_t = \sum_{i=1}^{n}\alpha_{ti}(h_iW_d^V), \ \ 
\alpha_{ti} = \frac{exp(\beta_{ti})}{\sum_{k=1}^{n} exp(\beta_{tk})}, \ \ \beta_{ti} = \frac{(q_{t}W_d^Q)(h_i{W_d^K}) ^T}{\sqrt{d}} + m^{\textit{buffer}}_{ti}
\end{equation}

The other two regular attention heads $z_t$ are computed as originally described in \citet{Vaswani2017}. All resulting attention vectors are combined and passed through subsequent linear and \textsc{Softmax} layers (as depicted in Fig.~\ref{fig:network}) to ultimately select the next action $a_t$ from the permitted transitions in state configuration $c_t$, according to a specific transition system $S$.

Finally, note that this neural architecture is flexible enough to implement not only the transition systems described in Section~\ref{sec:shift-reduce}, but also any shift-reduce parser for task-oriented semantic parsing.

\paragraph{Training objective} 
Each shift-reduce parser 
is trained through the minimization of the overall log loss (implemented as a cross-entropy loss) when selecting the correct sequence of transitions
$A = a_0, \dots , a_{m-1}$ to generate the gold TOP tree $Y_g$ for the user utterance $X$:
\begin{equation}
   \mathcal{L}(\theta) = - \sum_{t=0}^{T} log P_\theta (a_t | a_{<t},X) 
\end{equation}
\noindent where the transition $a_t$ (predicted in time step $t$) is conditioned by previous action predictions ($a_{<t}$).

\section{Experiments}
\label{sec:experiments}
\subsection{Setup}
\paragraph{Data} 
We conduct experiments on the main benchmark for task-oriented semantic parsing of compositional queries: the Facebook TOP datasets. The initial version (\texttt{TOP})\footnote{\url{http://fb.me/semanticparsingdialog}} was introduced by \citet{gupta-etal-2018-semantic-parsing}, who annotated utterances with multiple nested intents across two domains: \textit{event} and \textit{navigation}. This was further extended by \citet{chen-etal-2020-low} in the second version (\texttt{TOPv2})\footnote{\url{https://fb.me/TOPv2Dataset}}, which added six additional domains: \textit{alarm}, \textit{messaging}, \textit{music}, \textit{reminder}, \textit{timer} and \textit{weather}. While the first version presents user queries with a high degree of compositionality, the extension \texttt{TOPv2} introduced some domains (such as \textit{music} and \textit{weather}) where all utterances can be parsed with flat trees. Table~\ref{tab:treebanks} provides some statistics of the \texttt{TOP} and \texttt{TOPv2} datasets.

\begin{table}[h]
\centering
\begin{tabular}{@{\hskip 0pt}llcccccc@{\hskip 0pt}}
\toprule
\textbf{Dataset} & \textbf{Domain} & \textbf{Training} & \textbf{Valid} & \textbf{Test} & \textbf{Intents} & \textbf{Slots} & \textbf{\%Compos} \\
\midrule
\texttt{TOP} & event & 9,170 & 1,336 & 2,654 & 11 & 17 & 20\% \\
& navigation & 20,998 & 2,971 & 6,075 & 17 & 33 & 43\% \\
\midrule
\texttt{TOPv2} & alarm & 20,430 & 2,935 & 7,123 & 8 & 9 & 16\% \\
& messaging & 10,018 & 1,536 & 3,048 & 12 & 27 & 16\% \\
& music & 11,563 & 1,573 & 4,184 & 15 & 9 & 0\% \\
& reminder & 17,840 & 2,526 & 5,767 & 19 & 32 & 21\% \\
& timer & 11,524 & 1,616 & 4,252 & 11 & 5 & 4\% \\
& weather & 23,054 & 2,667 & 5,682 & 7 & 11 & 0\% \\
\bottomrule
\end{tabular}
\setlength{\abovecaptionskip}{4pt}
\caption{Data statistics for the Facebook TOP benchmark. We provide the number of queries in the training, validation and test splits; along with the number of intents and slots. Additionally, we include the percentage of compositional queries (\textit{i.e.,} utterances parsed by non-flat trees with depth $>$ 2).}
\label{tab:treebanks}
\end{table}

Furthermore, \texttt{TOPv2} offers specific splits designed to evaluate task-oriented semantic parsers in a \textit{low-resource domain adaptation} scenario. The conventional approach involves utilizing some samples from the \textit{reminder} and \textit{weather} domains as target domains, while considering the remaining six full domains (including \textit{event} and \textit{navigation} from \texttt{TOP}) as source domains if necessary. Moreover, instead of selecting a fixed number of training samples per target domain, \texttt{TOPv2} adopts a SPIS (samples per intent and slot) strategy. For example, a 25 SPIS strategy entails randomly selecting the necessary number of samples to ensure at least 25 training instances for each intent and slot of the target domain. To facilitate a fair comparison, we evaluate our approach on the training, test and validation splits at both 25 SPIS and 500 SPIS for the target domains \textit{reminder} and \textit{weather}, as provided in \texttt{TOPv2}. Additionally, following the methodology proposed by \citet{chen-etal-2020-low}, we employ a joint training strategy in the 25 SPIS setting, wherein the training data from the source domain is combined with the training split from the target domain.

Finally, we further evaluate our shift-reduce parsers on a variant of the \texttt{TOPv2} dataset (referred to as \texttt{TOPv2$^*$}). This variant comprises domains with a high percentage of hierarchical structures: \textit{alarm}, \textit{messaging} and \textit{reminder}. Our aim is to rigorously test the three proposed transition systems on complex compositional queries, excluding those domains that can be fully parsed with flat trees, which are more easily handled by traditional slot-filling methods.

\paragraph{Evaluation} 
We use the official TOP scoring script for performance evaluation, which reports three different metrics:
\begin{itemize}
    \item \textit{Exact match} accuracy (EM), which measures the percentage of full trees correctly built.
    \item \textit{Labeled bracketing} F$_1$ score (F$_1$), which compares the non-terminal label and span of each predicted constituent against the gold standard. This is similar to the scoring method provided by the EVALB script\footnote{\url{https://nlp.cs.nyu.edu/evalb/}} for constituency parsing \citep{black91}, but it also includes pre-terminal nodes in the evaluation.
    \item \textit{Tree-labeled} F$_1$ score (TF$_1$), which evaluates the subtree structure of each predicted constituent against the gold tree.
\end{itemize}
Recent research often reports only the EM accuracy; however, in line with \citet{gupta-etal-2018-semantic-parsing}, we also include F$_1$ and TF$_1$ scores to provide a more comprehensive comparison of the proposed transition systems. Lastly, for each experiment, we present the average score and standard deviation across three runs with random initialization.

\paragraph{Implementation details} 
Our neural architecture was built upon the Stack-Transformer framework developed by \citet{fernandez-astudillo-etal-2020-transition} using the FAIRSEQ toolkit \citep{fairseq}. We maintained consistent hyperparameters across all experiments, based on those specified by \citet{fernandez-astudillo-etal-2020-transition}, with minor adjustments. Specifically, we used the Adam optimizer \citep{Adam} with $\beta_1 = 0.9$ and $\beta_2 = 0.98$, and a batch size of 3584 tokens. The learning rate was linearly increased for the first 4,000 training steps from 1$e^{-7}$ to 5$e^{-4}$, followed by a decrease using the \texttt{inverse-sqrt} scheduling scheme, with a minimum of 1$e^{-9}$ \citep{Vaswani2017}. Additionally, we applied a label smoothing rate of 0.01, a dropout rate of 0.3, and trained for 90 epochs. Furthermore, we averaged the weights from the three best checkpoints based on the validation split using greedy decoding, and employed a beam size of 10 for evaluation on the test set. All models were trained and tested on a single Nvidia TESLA P40 GPU with 24 GB of memory.

\paragraph{Baselines}
\new{In addition to evaluating the three proposed transition systems, we compare them against the leading shift-reduce parser for task-oriented dialogue: the system developed by \citet{Einolghozati2019ImprovingSP}. This model builds upon the system by \citet{gupta-etal-2018-semantic-parsing}, which uses a top-down transition system and a Stack-LSTM-based architecture, and enhances it with ELMo-based word embeddings, a majority-vote ensemble of seven parsers, and an SVM language model ranker. 
We also include current top-performing sequence-to-sequence models in our comparison \citep{rogali2020,aghajanyan-etal-2020-conversational,zhu-etal-2020-dont,shrivastava-etal-2021-span-pointer,shrivastava-etal-2023-retrieve}. For low-resource domain adaptation, we compare our models with the enhanced implementation by \citet{chen-etal-2020-low}, which is based on \citet{rogali2020} and specifically tested on the low-resource \texttt{TOPv2} splits. 
 Lastly, we incorporate the recent state-of-the-art approach by \citet{do-etal-2023-structsp}, which employs a language model enhanced with semantic structured information, into both high-resource and low-resource comparisons.}

\begin{table*}[tbp]
\small
\centering
\begin{tabular}{@{\hskip 2pt}lc@{\hskip 0pt}c@{\hskip 5pt}c@{\hskip 10pt}c@{\hskip 5pt}c@{\hskip 5pt}c@{\hskip 2pt}}
\toprule
& \multicolumn{3}{c}{\texttt{TOP}} & \multicolumn{3}{c}{\texttt{TOPv2$^*$}} \\ 
\midrule
\textbf{Transition system}  & \textbf{EM}
&\textbf{F$_1$}
&\textbf{TF$_1$}
& \textbf{EM}
&\textbf{F$_1$}
&\textbf{TF$_1$}
\\
\midrule
\textbf{top-down}  & 86.66\tiny{$\pm$0.06} & 95.33\tiny{$\pm$0.07} & 90.80\tiny{$\pm$0.01} & 87.98\tiny{$\pm$0.09} & 94.51\tiny{$\pm$0.03} & 90.99\tiny{$\pm$0.09} \\
\textbf{bottom-up}  & 85.89\tiny{$\pm$0.10} & 94.86\tiny{$\pm$0.06} & 90.33\tiny{$\pm$0.05} & 86.27\tiny{$\pm$0.03} & 93.56\tiny{$\pm$0.06} & 89.57\tiny{$\pm$0.02}\\
\textbf{in-order}  & \textbf{87.15}\tiny{$\pm$0.01} & \textbf{95.57}\tiny{$\pm$0.15} & \textbf{91.18}\tiny{$\pm$0.13} & \textbf{88.11}\tiny{$\pm$0.07} & \textbf{94.60}\tiny{$\pm$0.04} & \textbf{91.11}\tiny{$\pm$0.07} \\
\bottomrule
\end{tabular}
\centering
\setlength{\abovecaptionskip}{4pt}
\caption{
Average performance across 3 runs on \texttt{TOP} and \texttt{TOPv2$^*$} test splits. Standard deviations 
are reported with $\pm$.
}
\label{tab:results1}
\end{table*}

\subsection{Results}
\paragraph{High-resource setting} 
We first present the evaluation results of the three described transition systems with Stack-Transformers on the \texttt{TOP} and \texttt{TOPv2$^*$} datasets in Table~\ref{tab:results1}. Regardless of the metric, the in-order algorithm consistently outperforms the other two alternatives on both datasets. Although the \texttt{TOP} dataset contains a higher percentage of compositional queries than \texttt{TOPv2$^*$}, the in-order parser shows a more significant accuracy advantage over the top-down parser on \texttt{TOP} (0.49 EM accuracy points) compared to \texttt{TOPv2$^*$} (0.13 EM accuracy points). The bottom-up approach notably underperforms compared to the other transition systems on both datasets.

\new{In Table~\ref{tab:sota}, we compare our shift-reduce parsers to strong baselines on the \texttt{TOP} dataset. Using frozen RoBERTa-based word embeddings, the in-order shift-reduce parser outperforms all existing methods under similar conditions, including sequence-to-sequence models that fine-tune language models for task-oriented parsing. Specifically, it surpasses the single-model and ensemble variants of the shift-reduce parser by \citet{Einolghozati2019ImprovingSP} by 3.22 and 0.89 EM accuracy points, respectively. Additionally, our best transition system achieves improvements of 0.41 and 0.05 EM accuracy points over top-performing sequence-to-sequence baselines initialized with RoBERTa \citep{zhu-etal-2020-dont} and BART \citep{aghajanyan-etal-2020-conversational}, respectively. The exceptions are the enhanced variant of \citet{Einolghozati2019ImprovingSP} (which uses an ensemble of seven parsers and an SVM language model ranker) and the two-staged system by \citet{do-etal-2023-structsp} that employs an augmented language model with hierarchical information, achieving the best accuracy to date on the \texttt{TOP} dataset.}

Lastly, our top-down parser with Stack-Transformers achieves accuracy comparable to the strongest sequence-to-sequence models using RoBERTa-based encoders \citep{rogali2020,zhu-etal-2020-dont}, and surpasses the single-model top-down shift-reduce baseline by \citet{Einolghozati2019ImprovingSP} by a wide margin (2.73 EM accuracy points).

\begin{table}[tbp]
\centering
\begin{tabular}{@{\hskip 2pt}lc@{\hskip 2pt}}
\toprule
\textbf{Parser} &\textbf{EM} \\
\midrule

\textit{\footnotesize (sequence-to-sequence models)} &   \\
\citet{rogali2020} + RoBERTa$_\textsc{fine-tuned}$ & 86.67 \\
\citet{aghajanyan-etal-2020-conversational} + RoBERTa$_\textsc{fine-tuned}$ & 84.52 \\
\citet{aghajanyan-etal-2020-conversational} + BART$_\textsc{fine-tuned}$ & 87.10 \\
\citet{zhu-etal-2020-dont} + RoBERTa$_\textsc{fine-tuned}$ & 86.74 \\
\citet{shrivastava-etal-2021-span-pointer} + RoBERTa$_\textsc{fine-tuned}$ & 85.07 \\
\citet{oh-etal-2022-improving} + BERT$_\textsc{fine-tuned}$ & 86.00 \\
\citet{shrivastava-etal-2023-retrieve} + RoBERTa$_\textsc{fine-tuned}$  & 86.14 \\
\hdashline[1pt/3pt]
\textit{\footnotesize (shift-reduce models)} &   \\
\citet{Einolghozati2019ImprovingSP} + ELMo  &  83.93 \\
\textbf{top-down shift-reduce parser} + RoBERTa & 86.66\\
\textbf{bottom-up shift-reduce parser} + RoBERTa &  85.89\\
\textbf{in-order shift-reduce parser} + RoBERTa &  \textbf{87.15}\\
\midrule
\midrule
\citet{Einolghozati2019ImprovingSP}  + ELMo + ensemble & 86.26  \\
\citet{Einolghozati2019ImprovingSP}  + ELMo + ensemble + SVM-Rank & 87.25  \\
\citet{do-etal-2023-structsp} + RoBERTa$_\textsc{fine-tuned}^\textsc{+ hierarchical information}$   & \textbf{88.18} \\
\bottomrule
\end{tabular}
\centering
\setlength{\abovecaptionskip}{4pt}
\caption{Comparison of exact match performance among state-of-the-art task-oriented parsers on the \texttt{TOP} test set. The first block encompasses sequence-to-sequence models, while the second block comprises shift-reduce parsers. In the last block, we additionally present the results of \citet{Einolghozati2019ImprovingSP} with ensembling (\textit{+ ensemble}) and language model re-ranking (\textit{+ SVM-Rank}), along with a novel approach that fine-tunes a standard RoBERTa language model by integrating additional semantic structured information (\textsc{+ hierarchical information}). Lastly, we denote with \textsc{fine-tuned} those approaches that utilize pre-trained language models directly as encoders and undergo fine-tuning for adaptation to task-oriented parsing.}
\label{tab:sota}
\end{table}

\begin{table}[tbp]
\centering
\begin{tabular}{@{\hskip 0pt}l@{\hskip 0pt}c@{\hskip 0pt}c@{\hskip 0pt}c@{\hskip 0pt}c@{\hskip 0pt}}
\toprule
 & \multicolumn{2}{c}{\textit{reminder}} & \multicolumn{2}{c}{\textit{weather}} \\
\textbf{Parser} & \textbf{25 SPIS} & \textbf{500 SPIS} & \textbf{25 SPIS} & \textbf{500 SPIS} \\
\midrule

\textit{\footnotesize (sequence-to-sequence models)} &   \\
\citet{chen-etal-2020-low} + RoBERTa$_\textsc{fine-tuned}$ & - & 71.9 & - & 83.5 \\
\citet{chen-etal-2020-low} + BART$_\textsc{fine-tuned}$ & 57.1 & 71.9 & 71.0 & 84.9 \\
\hdashline[1pt/3pt]
\textit{\footnotesize (shift-reduce models)} &   \\
\textbf{top-down S-R parser} + RoBERTa & \hphantom{0000}57.39\tiny{$\pm$0.27} & \hphantom{0000}\textbf{79.79}\tiny{$\pm$0.19} & \hphantom{0000}71.22\tiny{$\pm$1.03} & \hphantom{0000}83.19\tiny{$\pm$0.20}\\
\textbf{bottom-up S-R parser} + RoBERTa & \hphantom{0000}40.45\tiny{$\pm$0.88} & \hphantom{0000}68.65\tiny{$\pm$0.39} & \hphantom{0000}68.58\tiny{$\pm$0.99} & \hphantom{0000}74.13\tiny{$\pm$0.35}\\
\textbf{in-order S-R parser} + RoBERTa & \hphantom{0000}\textbf{60.56}\tiny{$\pm$0.12} & \hphantom{0000}\textbf{79.79}\tiny{$\pm$0.27} & \hphantom{0000}\textbf{73.36}\tiny{$\pm$0.08} & \hphantom{0000}\textbf{85.44}\tiny{$\pm$0.21}\\
\midrule
\midrule
\citet{do-etal-2023-structsp}+RoBERTa$_\textsc{fine-tuned}^\textsc{+hierar. inform.}$   & \textbf{72.12} & \textbf{82.28} & \textbf{77.96} & \textbf{88.08} \\
\bottomrule
\end{tabular}
\centering
\setlength{\abovecaptionskip}{4pt}
\caption{Comparison of exact match performance among top-performing task-oriented parsers on the test splits of \textit{reminder} and \textit{weather} domains within a low-resource setting. The first block compiles sequence-to-sequence models, while the second block encompasses shift-reduce (S-R) parsers. In the last block, we additionally present the results of the novel approach by \citet{do-etal-2023-structsp}, which involves fine-tuning a standard RoBERTa language model by integrating additional semantic structured information (\textsc{+ hierar. inform.}). Lastly, we denote with \textsc{fine-tuned} those approaches that utilize pre-trained language models directly as encoders and undergo fine-tuning for adaptation to task-oriented parsing.}
\label{tab:sota2}
\end{table}

\paragraph{Low-resource setting} 
Table~\ref{tab:sota2} presents the performance of our approach on low-resource domain adaptation. Across all SPIS settings, the in-order strategy consistently achieves the highest scores, not only among shift-reduce parsers but also compared to top-performing sequence-to-sequence models. Specifically, the in-order algorithm outperforms the BART-based sequence-to-sequence model by 3.5 and 2.4 EM accuracy points in the 25 SPIS setting of the \textit{reminder} and \textit{weather} domains, respectively. In the 500 SPIS setting, our best shift-reduce parser achieves accuracy gains of 7.9 and 0.5 EM points on the \textit{reminder} and \textit{weather} domains over the strongest sequence-to-sequence baseline. Notably, while the \textit{reminder} domain poses greater challenges due to the presence of compositional queries, our approach exhibits higher performance improvements in this domain compared to the \textit{weather} domain, which exclusively contains flat queries. Additionally, we include the state-of-the-art scores achieved by the system developed by \citet{do-etal-2023-structsp} by incorporating semantic structured information into the language model fine-tuning.

\paragraph{Discussion} 
Overall, our top-down and in-order shift-reduce parsers deliver competitive accuracies on the main Facebook TOP benchmark, surpassing the state of the art in both high-resource and low-resource settings in most cases. Furthermore, shift-reduce parsers ensure that the resulting structure is a well-formed tree in any setting, whereas sequence-to-sequence models may produce invalid trees due to the absence of grammar constraints during parsing. For instance, \citet{rogali2020} reported that 2\% of generated trees for the \texttt{TOP} test split were not well-formed. Although \citet{chen-etal-2020-low} did not document this information, we anticipate a significant increase in invalid trees in the low-resource setting. Finally, it is worth mentioning that techniques such as ensembling, re-ranking, or fine-tuning pre-trained language models are orthogonal to our approach and, while they may consume more resources, they can be directly implemented to further enhance performance.

\begin{figure}
\begin{center}
\includegraphics[width=\columnwidth]{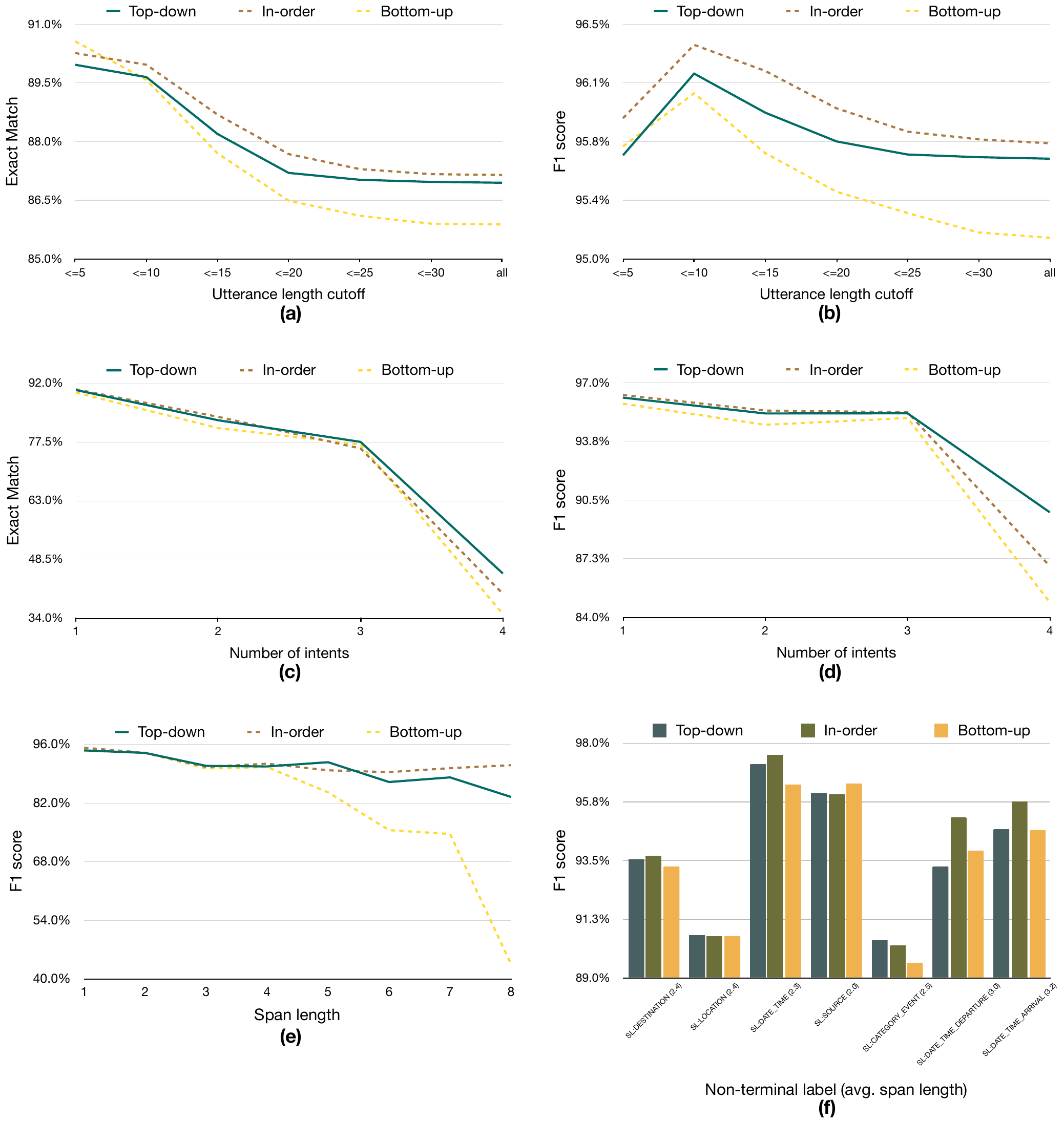}
\end{center}
\caption{Performance comparison of the three transition systems relative to utterance length and structural factors.}
\label{fig:analysis}
\end{figure}

\subsection{Analysis}
\label{sec:analysis}
To comprehend the variations in performance among the proposed transition systems, we conduct an error analysis focusing on utterance length and structural factors using the validation split of the \texttt{TOP} dataset.

 \paragraph{Utterance length} In Fig.~\ref{fig:analysis}(a) and (b), we present the EM and labeled bracketing F$_1$ scores achieved by each transition system across various utterance length cutoffs. It can be seen as the bottom-up algorithm yields higher EM accuracy for the shortest utterances ($\leq$ 5), but experiences a notable decline in accuracy for longer queries. While less pronounced than in the bottom-up strategy, both the in-order and top-down algorithms also exhibit a clear decrease in accuracy as utterance length increases. This outcome is anticipated as shift-reduce parsers are prone to \textit{error propagation}: earlier mistakes in the transition sequence can lead the parser into suboptimal state configurations, resulting in further erroneous decisions later on. Notably, the in-order approach consistently outperforms the top-down baseline across all length cutoffs.

 \paragraph{Query compositionality} Fig.s~\ref{fig:analysis}(c) and (d) depict the EM and labeled F$_1$ scores achieved by each algorithm on queries with varying numbers of intents per utterance. We observe that the in-order transition system attains higher EM accuracy on utterances with fewer than 3 intents. However, its performance on more complex queries is surpassed by the top-down approach. A similar trend is evident when evaluating performance using the labeled F$_1$ score: the top-down strategy outperforms the in-order algorithm on queries with 4 intents. While this might suggest that a purely top-down approach is preferable for processing utterances with a compositionality exceeding 3 intents, it is essential to note that the number of queries with 4 intents in the validation split is relatively low (just 20 utterances with 4 intents, compared to 2,525, 1,179, and 307 utterances with 1, 2, and 3 intents, respectively). Consequently, its impact on the overall performance is limited. Finally, both plots also indicate that the bottom-up approach consistently underperforms the other two alternatives, except on queries with 3 intents, where it surpasses the in-order strategy in EM accuracy.
 

 \paragraph{Span length and non-terminal prediction} 
 Fig.~\ref{fig:analysis}(e) illustrates the performance achieved by each transition system on span identification relative to different lengths, while Fig.~\ref{fig:analysis}(f) demonstrates the accuracy obtained by each algorithm on labeling constituents with the most frequent non-terminals (including the average span length in brackets). In Fig.~\ref{fig:analysis}(e), we observe that error propagation affects span identification, as accuracy decreases on longer spans, which require a longer transition sequence to be constructed and are thus more susceptible to error propagation. Additionally, the bottom-up transition system exhibits significant accuracy losses in producing constituents with longer spans. This can be attributed to the fact that, while the other two alternatives use a \textsc{Non-Terminal-}\textit{L} action to mark the beginning of the future constituent, the bottom-up strategy determines the entire span with a single \textsc{Reduce\#k-}\textit{L} transition at the end of the constituent creation. This approach, being more susceptible to error propagation, struggles with \textsc{Reduce\#k-}\textit{L} transitions with higher $k$ values, which are less frequent in the training data and hence more challenging to learn. Regarding the in-order algorithm, it appears to be more robust than the top-down transition system on constituents with the longest span, indicating that the advantages of the in-order strategy (explained in Section~\ref{sec:inorder}) mitigate the impact of error propagation.
Moreover, in Fig.~\ref{fig:analysis}(f), we observe that the in-order parser outperforms the other methods in predicting frequent non-terminal labels in nearly all cases, resulting in significant differences in accuracy in building slot constituents with the longest span, such as \texttt{SL:DATE\_TIME\_DEPARTURE} and \texttt{SL:DATE\_TIME\_ARRIVAL}. The only exceptions are slot constituents \texttt{SL:SOURCE} and \texttt{SL:CATEGORY\_EVENT}, where the bottom-up and top-down algorithms respectively achieve higher accuracy.

\section{Conclusions}
\label{sec:conclusion}
In this paper, we introduce innovative shift-reduce semantic parsers tailored for processing task-oriented dialogue utterances. In addition to the commonly used top-down algorithm for this task, we adapt the bottom-up and in-order transition systems from constituency parsing to generate well-formed TOP trees. Moreover, we devise a more robust neural architecture that, unlike previous shift-reduce approaches, leverages Stack-Transformers and RoBERTa-based contextualized word embeddings.

We extensively evaluate the three proposed algorithms across high-resource and low-resource settings, as well as multiple domains of the widely-used Facebook TOP benchmark. This marks the first evaluation of a shift-reduce approach in low-resource task-oriented parsing, to the best of our knowledge. Through these experiments, we demonstrate that the in-order transition system emerges as the most accurate alternative, surpassing all existing shift-reduce parsers not enhanced with re-ranking. Furthermore, it advances the state of the art in both high-resource and low-resource settings, surpassing all top-performing sequence-to-sequence baselines, including those employing larger pre-trained language models like BART.

\new{Additionally, it is worth noting that our approach holds potential for further enhancement through techniques such as ensemble parsing with a ranker, as developed by \citet{Einolghozati2019ImprovingSP}, or by specifically fine-tuning a RoBERTa-based encoder for task-oriented semantic parsing, as employed by the strongest sequence-to-sequence models. Finally, incorporating hierarchical semantic information, as successfully implemented by \citet{do-etal-2023-structsp}, is another avenue for improving our approach.}

\backmatter

\section*{Declarations}
\subsection*{Data availability}
The TOP and TOPv2 datasets used during the current research work are available in the Facebook repositories: \url{http://fb.me/semanticparsingdialog} and \url{https://fb.me/TOPv2Dataset}.

\subsection*{Fundings}
We acknowledge the European Research Council (ERC), which has funded this research under the European Union's Horizon 2020 research and innovation programme (FASTPARSE, grant agreement No 714150), ERDF/MICINN-AEI (PID2020-113230RB-C21, PID2020-113230RB-C22 and PID2023-147129OB-C22), Xunta de Galicia (ED431C 2020/11), and Centro de Investigación de Galicia ``CITIC'', funded by Xunta de Galicia and the European Union (ERDF - Galicia 2014-2020 Program), by grant ED431G 2019/01. Open Access funding provided thanks to the CRUE-CSIC agreement with Springer Nature.

\subsection*{Conflict of interest/Competing interests}
The authors declare that they have no known competing financial interests or personal relationships that could have appeared to influence the work reported in this article.

\bibliography{anthology,main}

\end{document}